\documentclass[10pt,twocolumn,letterpaper]{article}

\usepackage{cvpr}              

\usepackage{multirow}
\usepackage{array}
\usepackage{enumitem}
\usepackage{float}
\usepackage{dblfloatfix}
\usepackage[section]{placeins}

\definecolor{cvprblue}{rgb}{0.21,0.49,0.74}
\usepackage[pagebackref,breaklinks,colorlinks,allcolors=cvprblue]{hyperref}

\def\paperID{*****} 
\def\confName{CVPR}
\def\confYear{2026}

\title{GTF: Omnidirectional EPI Transformer for Light Field Super-Resolution}

\author{
Kunyu Li$^{1}$\quad Fei Wang$^{2}$\thanks{Corresponding authors.}\quad Lichao Zhang$^{3}$\footnotemark[1]\\
Junjie Liu$^{1}$\quad Bihong Li$^{2}$\\
{\large $^{1}$School of Computer Science and Technology, Xi'an Shiyou University}\\
{\large $^{2}$School of Electronic Engineering, Xi'an Shiyou University}\\
{\large $^{3}$Artificial Intelligence Research Institute, Shenzhen University of Advanced Technology}\\
{\tt\small kunyulee2002@gmail.com\quad 200102@xsyu.edu.cn\quad zhanglichao@suat-sz.edu.cn}\\
{\tt\small liujunie@foxmail.com\quad 3112735122@qq.com}\\
}

\begin{document}
\maketitle
\begin{abstract}
Light field (LF) image super-resolution benefits from Epipolar Plane Images (EPIs), whose line slopes explicitly encode disparity. However, existing Transformer-based LF SR methods mainly attend to horizontal and vertical EPIs, leaving diagonal epipolar geometry underexplored. We present \textbf{GTF}, an omnidirectional EPI Transformer that explicitly models horizontal, vertical, 45$^\circ$, and 135$^\circ$ EPIs within a unified reconstruction framework. GTF combines directional EPI processing, MacPI-based prior injection, adaptive directional fusion, and a topology-preserving feed-forward network to better exploit LF geometry. For the NTIRE 2026 fidelity tracks, we use GTF as the main model, while a lightweight \textbf{GTF-Tiny} variant targets the efficiency track. On five standard LF SR benchmarks covering both real-captured and synthetic scenes, GTF reaches 32.78~dB without inference-time enhancement, and stronger inference settings with EPSW and test-time augmentation further improve performance. Under the NTIRE 2026 efficiency constraint, GTF-Tiny attains 32.57~dB with only 0.915M parameters and 19.81 GFLOPs. In the NTIRE 2026 Light Field Image Super-Resolution Challenge, our submissions rank \textbf{3rd on Track~1 and Track~3 and 4th on Track~2}. Architecture-evolution, channel-width, and inference analyses further support the effectiveness of diagonal EPI modeling, directional fusion, and the lightweight design.
\end{abstract}

\section{Introduction}
\label{sec:intro}

Light field (LF) imaging captures the four-dimensional (4D) light distribution $\mathcal{L}(u, v, h, w)$, where $(u, v)$ denotes angular coordinates and $(h, w)$ denotes spatial coordinates~\cite{levoy1996lightfield,ng2005lightfield}. This representation enables applications such as depth estimation, refocusing, and view synthesis~\cite{kalantari2016learning}. However, the trade-off between angular and spatial resolution in commercial LF cameras motivates LF image super-resolution (SR). A key geometric structure in LF data is the Epipolar Plane Image (EPI), formed by fixing one angular and one spatial dimension. The slope of lines in an EPI directly encodes scene disparity~\cite{shin2018epinet}, which motivates EPI-aware LF reconstruction.

Early CNN-based methods explored diverse ways to exploit LF structure, including spatial-angular interaction~\cite{wang2020lfinternet}, disentangled modeling~\cite{wang2022distgssr}, and multi-orientation EPI processing~\cite{duong2023hlfsr}. Among them, HLFSR~\cite{duong2023hlfsr} is particularly relevant because it indicates the value of four EPI orientations, but still relies on shared convolutional filters rather than direction-specific long-range modeling. Recent Transformer-based models further improve LF SR by modeling broader dependencies. EPIT~\cite{liang2023epit} applies self-attention to horizontal and vertical EPIs and establishes a strong EPI-Transformer baseline, while a recent analysis further studies EPI-oriented Transformers for LF SR and disparity estimation~\cite{liang2026diving}. LF-DET~\cite{cong2023lfdet} instead emphasizes efficient spatial-angular token interaction and is not designed around explicit four-direction epipolar modeling. Hybrid Mamba/Transformer methods have also appeared: LFTransMamba~\cite{jin2025lftransmamba} relies on SAI and MacPI representations without explicit EPI attention, whereas LFTramba~\cite{liu2025lftramba} incorporates EPI Transformers but remains limited to the horizontal and vertical directions.

Despite this progress, diagonal EPIs remain largely absent from Transformer-based LF SR. This omission matters because diagonal views can provide disparity cues that are not always aligned with the horizontal or vertical axes. HLFSR~\cite{duong2023hlfsr} already suggests that these cues are useful, but its shared-weight CNN design cannot adapt the feature extractor to different EPI orientations and does not benefit from self-attention over long epipolar structures. In contrast, existing Transformer-based LF SR methods either focus on axis-aligned EPIs (EPIT, LFTramba) or prioritize efficient spatial-angular interaction without explicitly extending epipolar reasoning to diagonal directions (LF-DET). A gap therefore remains between multi-orientation LF geometry and modern attention-based LF reconstruction.

To address this gap, we propose \textbf{GTF}, an omnidirectional EPI Transformer for LF SR. GTF explicitly models horizontal, vertical, and diagonal EPIs, and fuses them through an adaptive directional fusion module. We further use a topology-preserving feed-forward network to better preserve 2D EPI structure, and develop a lightweight variant for the efficiency-constrained setting. Beyond standard benchmark evaluation on both real-captured and synthetic LF datasets, we also validate the design in the NTIRE 2026 Light Field Image Super-Resolution Challenge, where our submissions place 3rd on Track~1 and Track~3 and 4th on Track~2.

Our main contributions are:
\begin{itemize}[nosep,leftmargin=*]
    \item We introduce \textbf{GTF}, a Transformer-based LF SR framework that explicitly models horizontal, vertical, 45$^\circ$, and 135$^\circ$ EPIs for omnidirectional epipolar reasoning.
    \item We propose adaptive directional fusion and a topology-preserving FFN to improve angular-spatial representation learning within the GTF framework.
    \item We validate the proposed GTF model together with a lightweight \textbf{GTF-Tiny} variant (0.915M parameters, 19.81 GFLOPs) on standard benchmarks, through architecture evolution and channel-width analyses, and in the NTIRE 2026 challenge.
\end{itemize}

\section{Related Work}
\label{sec:related}

\subsection{Light Field Image Super-Resolution}

LF image SR aims to recover high-resolution sub-aperture images from their low-resolution counterparts while preserving the angular consistency of the 4D light field. Building on advances in single-image SR~\cite{lim2017edsr,zhang2018rcan,liang2021swinir}, CNN-based LF methods have explored various strategies to exploit spatial-angular structure. Yeung~\etal~\cite{yeung2018lfssr} proposed combinatorial geometry embedding for spatial SR. LF-InterNet~\cite{wang2020lfinternet} models spatial-angular interactions through alternating convolutions. LF-DFnet~\cite{wang2021lfdfnet} introduces deformable convolutions to handle non-uniform disparity. DistgSSR~\cite{wang2022distgssr} disentangles LF features into spatial and angular components for separate processing. LFASIN~\cite{chen2022lfasin} models angular and spatial interactions jointly. LFT~\cite{wang2023lft} introduces Transformer-based modeling into LF SR. HLFSR~\cite{duong2023hlfsr} is the closest CNN counterpart to our work: it processes four EPI orientations (0$^\circ$, 90$^\circ$, 45$^\circ$, 135$^\circ$), but uses shared convolutional weights rather than orientation-specific attention.

Transformer-based methods further improve LF SR by modeling long-range dependencies more explicitly. EPIT~\cite{liang2023epit} is the representative EPI-Transformer baseline, but its attention is restricted to horizontal and vertical EPIs. LF-DET~\cite{cong2023lfdet} pursues efficient spatial-angular correlation modeling and does not explicitly extend attention to diagonal EPIs. Ko~\etal~\cite{ko2023lfsr_transformer} further explore Transformer designs for LF reconstruction. Beyond Transformer-based models, IINet~\cite{liang2022lfiinet} studies intra-view and inter-view interaction for LF SR, but it likewise does not formulate explicit diagonal EPI attention. Overall, prior LF SR methods either remain axis-aligned in their epipolar reasoning or do not formulate diagonal EPI attention explicitly.

\subsection{EPI-Based Processing}

Epipolar Plane Images encode scene geometry through the slope of linear structures, which directly corresponds to disparity~\cite{shin2018epinet}. EPINet~\cite{shin2018epinet} first highlighted the value of multi-orientation EPI processing for LF depth estimation. ResLF~\cite{zhang2019reslf} extended EPI reasoning to LF SR, and HLFSR~\cite{duong2023hlfsr} further adopted four-directional EPI processing with shared convolutional weights.

In the Transformer paradigm, EPIT~\cite{liang2023epit} and follow-up methods show that self-attention is well suited to EPI modeling, but they focus on horizontal and vertical slices. Diagonal EPIs provide complementary 45$^\circ$ and 135$^\circ$ disparity cues, yet they remain absent from Transformer-based LF SR pipelines. GTF addresses this gap with direction-specific attention branches rather than shared directional filters.

\subsection{State Space Models for Vision}

Selective state space models (SSMs), especially Mamba~\cite{gu2023mamba}, have become attractive alternatives to Transformers for efficient long-range modeling. VMamba~\cite{liu2024vmamba} adapts this idea to vision, and recent LF SR methods explore related hybrids. LFTransMamba~\cite{jin2025lftransmamba} combines Mamba with Transformer components while operating on SAI and MacPI representations instead of explicit EPI attention. LFTramba~\cite{liu2025lftramba} is closer to our setting because it mixes Mamba with EPI Transformers, but its EPI modeling is still confined to horizontal and vertical directions. LFMix~\cite{yu2025lfmix} pursues lightweight hybrid modeling. These methods reinforce the value of efficient global modeling, but none introduces diagonal EPI attention into a Transformer-based LF SR framework.

\section{Proposed Method}
\label{sec:method}

\subsection{Overview}
\label{sec:overview}

\begin{figure*}[t]
\centering
\includegraphics[width=\linewidth,trim=10 6 10 12,clip]{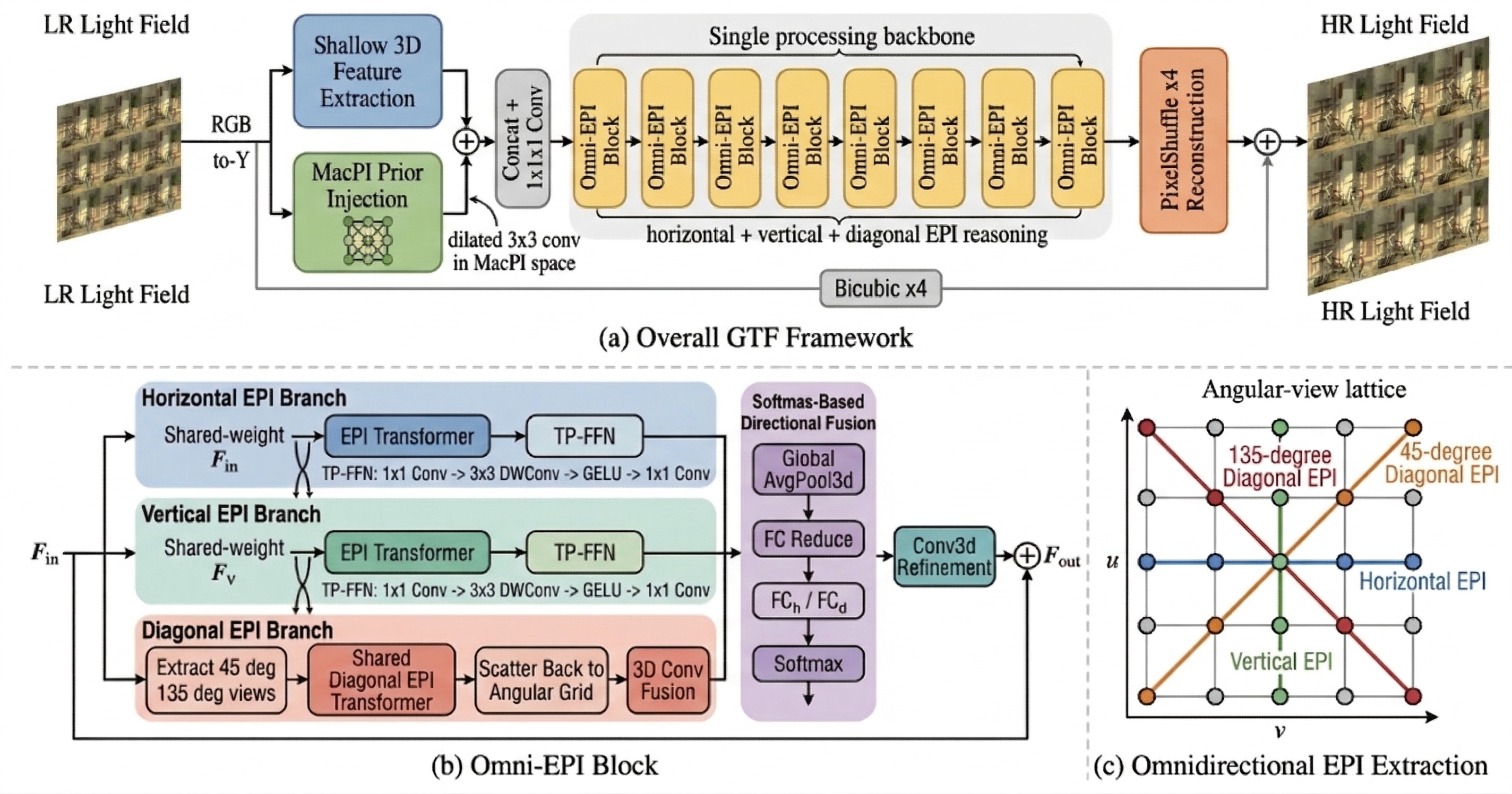}
\caption{Overview of the proposed GTF framework. (a) Overall network architecture of GTF with MacPI prior injection, stacked Omni-EPI Blocks, and residual reconstruction. (b) Architecture of the Omni-EPI Block, where horizontal, vertical, and diagonal branches are adaptively fused before residual refinement. (c) Illustration of omnidirectional EPI extraction from the angular grid.}
\label{fig:method_overview}
\end{figure*}

Given a low-resolution (LR) light field $\mathcal{L} \in \mathbb{R}^{U \times V \times H \times W}$ with angular dimensions $U = V = 5$ and spatial dimensions $H \times W$, our goal is to reconstruct a high-resolution (HR) light field with spatial size $\alpha H \times \alpha W$, where $\alpha = 4$ is the upsampling scale factor.

The overall pipeline of GTF is illustrated in Fig.~\ref{fig:method_overview}(a). Starting from the LR LF input, the fidelity model first extracts shallow volumetric features and in parallel injects a MacPI-based prior. The fused features are then processed by $N$ cascaded Omni-EPI Blocks that combine horizontal, vertical, and diagonal EPI reasoning with adaptive directional fusion. Finally, PixelShuffle upsampling~\cite{shi2016pixelshuffle} reconstructs the spatial resolution, and a global residual connection from the bicubic-upsampled input produces the final SR output.

For the fidelity setting, we use a trained GTF model as the main method in Track~1/3. It operates on RGB LF input, internally converts it to YCbCr, reconstructs the luminance component, and converts the final result back to RGB. For the efficiency setting, we use GTF-Tiny under the official NTIRE constraints. Both variants follow the same high-level principle of omnidirectional EPI modeling with directional fusion, while the lightweight model adopts a simplified implementation tailored to the efficiency budget.

\subsection{MacPI Prior Injection}
\label{sec:macpi}

The main fidelity model incorporates a MacPI-based prior inspired by HLFSR~\cite{duong2023hlfsr}. We rearrange the luminance component of the LR LF into a 2D macro-pixel image of size $UH \times VW$, apply a dilated $3\times3$ convolution with dilation factor $U$, and rearrange the result back to the 3D LF feature format. The prior feature is then fused with the shallow volumetric feature by channel concatenation followed by a $1\times1\times1$ convolution, providing angular-spatial cues before the omnidirectional EPI backbone.

\subsection{Omnidirectional EPI Processing}
\label{sec:altfilter}

Epipolar plane images (EPIs) are 2D slices of the light field that encode scene geometry through their slope patterns: the disparity of a scene point is directly proportional to the slope of its corresponding line in the EPI~\cite{shin2018epinet}.
Prior Transformer-based methods~\cite{liang2023epit,cong2023lfdet} focus on horizontal and vertical EPIs, leaving diagonal disparity information unmodeled. Recent analysis of epipolar Transformers for LF SR and disparity estimation further studies the role of EPI-oriented attention in LF processing~\cite{liang2026diving}.
In contrast, HLFSR~\cite{duong2023hlfsr} processes four EPI directions but shares CNN weights across them, which weakens direction-specific modeling.

Our Omni-EPI Block addresses both limitations by employing directional EPI branches for horizontal, vertical, and diagonal EPIs. The detailed structure is shown in Fig.~\ref{fig:method_overview}(b).

\subsubsection{EPI Transformer Branch}
\label{sec:basictrans}

Each directional branch follows the same high-level template: EPI tokens are rearranged from the 5D LF feature tensor, processed by self-attention, and refined by a topology-preserving feed-forward network:
\begin{align}
\mathbf{X}' &= \mathbf{X} + \gamma_1 \cdot \operatorname{DropPath}\!\left(\operatorname{MHSA}(\operatorname{LN}(\mathbf{X}))\right), \label{eq:attn} \\
\mathbf{Y}  &= \mathbf{X}' + \gamma_2 \cdot \operatorname{DropPath}\!\left(\operatorname{TP\!-\!FFN}(\operatorname{LN}(\mathbf{X}'))\right), \label{eq:ffn}
\end{align}
where $\mathbf{X}$, $\mathbf{X}'$, and $\mathbf{Y}$ denote the input, intermediate, and output token sequences of one directional branch, respectively; $\operatorname{LN}(\cdot)$ is LayerNorm~\cite{ba2016layernorm}; $\operatorname{MHSA}(\cdot)$ is multi-head self-attention~\cite{vaswani2017attention}; $\operatorname{DropPath}(\cdot)$ denotes stochastic depth on the residual branch; and $\gamma_1,\gamma_2$ are learnable LayerScale parameters. In the main fidelity model, this branch is implemented with multi-head self-attention, learnable LayerScale parameters, and local epipolar attention. The horizontal and vertical branches share the same EPI Transformer weights, while the diagonal branch uses the same branch template on diagonal EPI sequences. In the lightweight GTF-Tiny variant, this branch is simplified with scaled dot-product attention and no explicit local mask.

\subsubsection{Axis-Aligned EPI Branches}
\label{sec:hvbranch}

Horizontal and vertical EPIs provide the two standard axis-aligned epipolar views. Horizontal EPIs are formed by fixing the vertical angular index $v$ and the vertical spatial coordinate $y$, yielding 2D slices that capture horizontal disparity. We construct them by rearranging the 5D feature tensor as
\begin{equation}
\mathrm{b\;c\;(u\,v)\;h\;w} \;\rightarrow\; \mathrm{b\;c\;(v\,w)\;u\;h},
\end{equation}
producing horizontal EPI sequences over which self-attention captures cross-view disparity patterns. Analogously, vertical EPIs are obtained by fixing the horizontal angular index $u$ and the horizontal spatial coordinate $x$, with rearrangement
\begin{equation}
\mathrm{b\;c\;(u\,v)\;h\;w} \;\rightarrow\; \mathrm{b\;c\;(u\,h)\;v\;w},
\end{equation}
yielding vertical EPI sequences for vertical epipolar reasoning.

\subsubsection{Diagonal EPI Branch}
\label{sec:diabranch}

The diagonal EPI branch is the key extension of GTF. Horizontal and vertical EPIs capture only axis-aligned disparity patterns, whereas diagonal epipolar slices provide complementary cues for structures that are not well aligned to those axes. This complementarity can be useful for thin structures, oblique boundaries, and occlusion transitions, where disparity traces are not always well explained by only horizontal or vertical slices. We therefore explicitly extract and process diagonal EPIs from the angular grid, as illustrated in Fig.~\ref{fig:method_overview}(c).

From the $U \times V$ angular grid, we extract two sets of diagonal sub-aperture views:
\begin{itemize}[nosep,leftmargin=*]
\item \textbf{45$^\circ$ diagonal:} views at angular positions $(i, i)$ for $i = 0, \ldots, U{-}1$, forming a diagonal traversal from the top-left to the bottom-right of the angular grid.
\item \textbf{135$^\circ$ diagonal:} views at angular positions $(i, U{-}1{-}i)$ for $i = 0, \ldots, U{-}1$, traversing from the top-right to the bottom-left.
\end{itemize}
Each diagonal yields $U$ views, which are rearranged into EPI sequences and processed with the same branch template used in the axis-aligned directions.

For reproducibility, the diagonal branch follows the tensor flow below. Starting from the 5D LF feature tensor $\mathbf{F} \in \mathbb{R}^{B \times C \times U^2 \times H \times W}$, we first reshape it into a 6D angular grid $\mathbb{R}^{B \times C \times U \times V \times H \times W}$. We then gather diagonal views with index sets $\{(i,i)\}_{i=0}^{U-1}$ and $\{(i,U-1-i)\}_{i=0}^{U-1}$. Each gathered tensor has shape $\mathbb{R}^{B \times C \times U \times H \times W}$ and is further rearranged to $\mathbb{R}^{B \times C \times W \times U \times H}$ so that the diagonal angular dimension and one spatial dimension form an EPI sequence for Transformer processing.

The diagonal attention outputs are scattered back to their original positions in the full $U \times V$ angular grid and refined with a 3D convolution:
\begin{equation}
\begin{split}
\mathbf{F}_d = \mathbf{F} + \operatorname{Conv3D}\!\Big(&\operatorname{Scatter}\!\big(\operatorname{Attn}_{45^\circ}(\mathbf{E}_{45^\circ}),\\
&\operatorname{Attn}_{135^\circ}(\mathbf{E}_{135^\circ})\big)\Big),
\end{split}
\label{eq:diag}
\end{equation}
where $\mathbf{E}_{45^\circ}$ and $\mathbf{E}_{135^\circ}$ denote the extracted diagonal EPI features, $\operatorname{Attn}_{45^\circ}(\cdot)$ and $\operatorname{Attn}_{135^\circ}(\cdot)$ denote the EPI Transformer branch applied to the $45^\circ$ and $135^\circ$ diagonal sequences, respectively, and $\operatorname{Scatter}(\cdot)$ places the processed features back into the full angular grid. Concretely, the scatter step initializes a zero tensor with the full angular layout, writes the processed $45^\circ$ and $135^\circ$ responses back to their original angular coordinates, and finally flattens the angular grid to the standard $\mathbb{R}^{B \times C \times U^2 \times H \times W}$ layout before the residual 3D refinement.
This design extends Transformer-based LF SR from two-direction to four-direction epipolar reasoning.

\subsubsection{Topology-Preserving FFN}
\label{sec:tpffn}

Standard Transformer architectures employ a point-wise feed-forward network (FFN) that operates independently on each token.
When applied to EPI sequences, this 1D operation ignores the inherent 2D topology of each slice.

To address this, we propose the Topology-Preserving FFN (TP-FFN), which temporarily restores the 2D structure of the EPI during feed-forward processing:
\begin{multline}
\operatorname{TP\!-\!FFN}(\mathbf{X}) = \operatorname{Conv}_{1 \times 1}^{\downarrow}\Big(\operatorname{GELU}\big(\operatorname{DWConv}_{3 \times 3}\\
\big(\operatorname{Conv}_{1 \times 1}^{\uparrow}(\operatorname{LN}(\hat{\mathbf{X}}))\big)\big)\Big),
\label{eq:tpffn}
\end{multline}
where $\hat{\mathbf{X}}$ is the 2D feature map reshaped from the token sequence $\mathbf{X}$, $\operatorname{Conv}_{1 \times 1}^{\uparrow}$ expands the channel dimension by a factor of $r$ (expansion ratio), $\operatorname{DWConv}_{3 \times 3}$ is a depthwise convolution~\cite{howard2017mobilenets} that preserves local spatial structure, and $\operatorname{Conv}_{1 \times 1}^{\downarrow}$ projects the feature back to the original channel dimension.
\begin{table*}[!t]
\centering
\setlength{\abovecaptionskip}{3pt}
\setlength{\belowcaptionskip}{0pt}
\caption{Quantitative comparison for $4\times$ LFSR with $5{\times}5$ angular resolution. PSNR/SSIM are reported on the Y channel. $\dagger$: EPSW+TTA. $\ddagger$: EPSW only. Best and second-best results are highlighted in bold and with underlining, respectively.}
\label{tab:sota}
\resizebox{\textwidth}{!}{%
\begin{tabular}{@{}lcccccccc@{}}
\toprule
Method & Params (M) & FLOPs (G) & EPFL & HCI\_new & HCI\_old & INRIA & STFgantry & Average \\
\midrule
Bicubic & --- & --- & 25.26/0.8324 & 27.72/0.8517 & 32.58/0.9344 & 26.95/0.8867 & 26.09/0.8452 & 27.72/0.8701 \\
EDSR~\cite{lim2017edsr} & 38.89 & 1017 & 27.84/0.8854 & 29.60/0.8869 & 35.18/0.9536 & 29.66/0.9257 & 28.70/0.9072 & 30.20/0.9118 \\
RCAN~\cite{zhang2018rcan} & 15.36 & 408.5 & 27.88/0.8863 & 29.63/0.8886 & 35.20/0.9548 & 29.76/0.9276 & 28.90/0.9131 & 30.27/0.9141 \\
resLF~\cite{zhang2019reslf} & 8.64 & 39.7 & 28.27/0.9035 & 30.73/0.9107 & 36.71/0.9682 & 30.34/0.9412 & 30.19/0.9372 & 31.25/0.9322 \\
LFSSR~\cite{yeung2018lfssr} & 1.77 & 128.4 & 28.27/0.9118 & 30.72/0.9145 & 36.70/0.9696 & 30.31/0.9467 & 30.15/0.9426 & 31.23/0.9370 \\
LF-ATO~\cite{jin2020lfato} & 1.36 & 687.0 & 28.52/0.9115 & 30.88/0.9135 & 37.00/0.9699 & 30.71/0.9484 & 30.61/0.9430 & 31.54/0.9373 \\
LF-InterNet~\cite{wang2020lfinternet} & 5.48 & 50.1 & 28.67/0.9162 & 30.98/0.9161 & 37.11/0.9716 & 30.64/0.9491 & 30.53/0.9409 & 31.58/0.9388 \\
MEG-Net~\cite{zhang2021megnet} & 1.77 & 102.2 & 28.74/0.9160 & 31.10/0.9177 & 37.28/0.9716 & 30.66/0.9490 & 30.77/0.9453 & 31.71/0.9399 \\
LF-DFnet~\cite{wang2021lfdfnet} & --- & --- & 28.77/0.9165 & 31.23/0.9196 & 37.32/0.9718 & 30.83/0.9503 & 31.15/0.9494 & 31.86/0.9415 \\
DPT~\cite{wang2022dpt} & 3.78 & 66.6 & 28.93/0.9170 & 31.19/0.9188 & 37.39/0.9721 & 30.96/0.9503 & 31.14/0.9488 & 31.93/0.9414 \\
IINet~\cite{liang2022lfiinet} & 4.88 & 57.4 & 29.11/0.9188 & 31.36/0.9208 & 37.62/0.9734 & 31.08/0.9515 & 31.21/0.9502 & 32.08/0.9429 \\
DistgSSR~\cite{wang2022distgssr} & 3.58 & 65.4 & 28.99/0.9195 & 31.38/0.9217 & 37.56/0.9732 & 30.99/0.9519 & 31.65/0.9535 & 32.11/0.9440 \\
LFT~\cite{wang2023lft} & 1.16 & 57.6 & 29.25/0.9210 & 31.46/0.9218 & 37.63/0.9735 & 31.20/0.9524 & 31.86/0.9548 & 32.28/0.9447 \\
HLFSR~\cite{duong2023hlfsr} & 13.87 & 182.5 & 29.20/0.9222 & 31.57/0.9238 & 37.78/0.9742 & 31.24/0.9534 & 31.64/0.9537 & 32.29/0.9455 \\
EPIT~\cite{liang2023epit} & 1.47 & 75.0 & 29.34/0.9197 & 31.51/0.9231 & 37.68/0.9737 & 31.37/0.9526 & 32.18/0.9571 & 32.40/0.9452 \\
LF-DET~\cite{cong2023lfdet} & 1.69 & 51.2 & 29.47/0.9230 & 31.56/0.9235 & 37.84/0.9744 & 31.39/0.9534 & 32.14/0.9573 & 32.48/0.9463 \\
\midrule
GTF-Tiny$^\ddagger$ & \textbf{0.915} & \textbf{19.8} & \underline{29.88}/\underline{0.9197} & 31.49/0.9222 & 37.67/0.9735 & \underline{32.08}/\underline{0.9524} & 31.75/0.9534 & 32.57/0.9442 \\
GTF & 16.26 & 335 & 29.82/0.9250 & \underline{31.75}/\underline{0.9253} & \underline{38.01}/\underline{0.9749} & 31.73/0.9554 & \underline{32.57}/\underline{0.9593} & \underline{32.78}/\underline{0.9480} \\
GTF$^\dagger$ & 16.26 & 335 & \textbf{30.39}/\textbf{0.9262} & \textbf{31.90}/\textbf{0.9267} & \textbf{38.16}/\textbf{0.9754} & \textbf{32.50}/\textbf{0.9563} & \textbf{32.84}/\textbf{0.9608} & \textbf{33.16}/\textbf{0.9491} \\
\bottomrule
\end{tabular}%
}
\end{table*}

The output is flattened back to the 1D token sequence. We use an expansion ratio of $r = 4$ in the default configuration. The depthwise convolution preserves 2D locality while complementing the global receptive field of self-attention.

\subsection{Adaptive Directional Fusion}
\label{sec:gating}

After the three directional branches produce $\mathbf{F}_h$, $\mathbf{F}_v$, and $\mathbf{F}_d$, we fuse them with an adaptive channel-wise directional fusion module so that the three epipolar directions are weighted according to the current feature content.

The fusion mechanism first computes a channel descriptor by global average pooling over the sum of branch outputs:
\begin{equation}
\mathbf{z} = \operatorname{AvgPool3D}\!\left(\mathbf{F}_h + \mathbf{F}_v + \mathbf{F}_d\right) \in \mathbb{R}^{C}.
\label{eq:gate_pool}
\end{equation}
The descriptor is then passed through a lightweight bottleneck network to generate directional weights. In the fidelity model, the bottleneck first reduces the channel dimension from $C$ to $C/\rho$ and then predicts one channel-wise weight tensor for each direction; in GTF-Tiny, this path is implemented with a two-layer MLP and sigmoid activation. Here, $\rho$ is the channel-reduction ratio, $\phi_1$ maps $\mathbb{R}^{C}$ to $\mathbb{R}^{C/\rho}$, and $\phi_2$ maps $\mathbb{R}^{C/\rho}$ to $\mathbb{R}^{3C}$. The gating path can be summarized as
\begin{equation}
[\mathbf{g}_h,\mathbf{g}_v,\mathbf{g}_d] = \phi_2\!\left(\phi_1(\mathbf{z})\right),
\label{eq:gate_mlp}
\end{equation}
where the output is reshaped into three channel-wise gates $\mathbf{g}_h$, $\mathbf{g}_v$, and $\mathbf{g}_d$ and broadcast over the spatial-angular dimensions. The directional weights are then applied to the three branches before residual fusion:
\begin{equation}
\mathbf{F}_{\text{out}} = \operatorname{Conv3D}\!\left(\mathbf{g}_h \odot \mathbf{F}_h + \mathbf{g}_v \odot \mathbf{F}_v + \mathbf{g}_d \odot \mathbf{F}_d\right) + \mathbf{F}_{\text{in}},
\label{eq:gate_fuse}
\end{equation}
where $\odot$ denotes channel-wise multiplication and $\mathbf{F}_{\text{in}}$ is the input to the Omni-EPI Block, forming a residual connection.

Compared with covariance-based angular modulation in HLFSR~\cite{duong2023hlfsr}, our fusion is learned directly from pooled directional responses. In the main fidelity model, this idea is instantiated as channel-wise directional fusion over the horizontal, vertical, and diagonal branches. In GTF-Tiny, we use a lighter two-layer MLP with sigmoid activation to produce three channel-wise weights before residual refinement.

\subsection{Lightweight GTF for Efficiency Track}
\label{sec:track2}

For the NTIRE 2026 efficiency track, GTF-Tiny is designed under the ${<}1$M parameter and ${<}20$G FLOP constraints by reducing model width and depth, simplifying attention, and adopting lighter aggregation:

\begin{itemize}[nosep,leftmargin=*]
\item \textbf{Reduced capacity:} channels $C\!: 128 \to 32$, blocks $N\!: 8 \to 6$, attention heads: $8 \to 4$, TP-FFN expansion ratio $r\!: 4 \to 2$.
\item \textbf{Simplified attention:} local masks are removed in favor of full scaled dot-product attention, reducing both computation and mask-construction overhead.
\item \textbf{Decoupled axis-aligned branches:} separate horizontal and vertical EPI Transformer branches are used, while a dedicated diagonal branch is retained for 45$^\circ$ and 135$^\circ$ processing.
\item \textbf{Angular position embedding:} a learnable embedding $\mathbf{P}_{\text{ang}} \in \mathbb{R}^{1 \times C \times U^2 \times 1 \times 1}$ is added to the initial features $\mathbf{F}_0$, providing explicit angular position information that compensates for the reduced model capacity.
\end{itemize}

To preserve both shallow structure and deeper context, GTF-Tiny further uses multi-level aggregation (MLA), which collects features from blocks $\{1, 3, 5\}$:
\begin{equation}
\mathbf{F}_{\text{mla}} = \operatorname{Conv3D}_{1 \times 1}\!\left([\mathbf{F}^{(1)},\; \mathbf{F}^{(3)},\; \mathbf{F}^{(5)}]\right),
\label{eq:mla}
\end{equation}
where $\mathbf{F}^{(1)}$, $\mathbf{F}^{(3)}$, and $\mathbf{F}^{(5)}$ denote the intermediate features collected after the 1st, 3rd, and 5th lightweight Omni-EPI Blocks, and $[\cdot,\cdot,\cdot]$ denotes channel concatenation. This concatenation produces a tensor of size $\mathbb{R}^{B \times 3C \times U^2 \times H \times W}$, and the pointwise $\operatorname{Conv3D}$ fuses it back to $C$ channels. Together with sigmoid-based channel-wise directional weighting after the three EPI branches, this hierarchical fusion helps retain fine-grained EPI structures. The main fidelity model instead uses a simpler deep residual stack over the Omni-EPI Blocks.

These modifications yield a model with 0.915M parameters and 19.81 GFLOPs, satisfying the efficiency constraints while maintaining competitive reconstruction quality, as demonstrated in Sec.~\ref{sec:experiments}.

\section{Training and Inference}
\label{sec:training}

\subsection{Loss Functions}

We compare two training losses: plain L1 and an Online Hard Example Mining (OHEM)~\cite{shrivastava2016ohem}-based objective. In our implementation, the OHEM variant combines hard-pixel selection with a pixel-wise Charbonnier-style robust loss and backpropagates only through the top-$k$ hardest pixels. We use $k=0.5$ for GTF and $k=0.8$ for GTF-Tiny; under our training setup, this choice improved validation performance, especially around difficult regions such as occlusion boundaries and thin structures.

\subsection{Training Protocol}

\paragraph{GTF.} For the fidelity setting, we use a trained GTF model as the main instantiation of the proposed framework. It is trained with Adam~\cite{kingma2015adam} using LR patch size $32 \times 32$, standard flip/rotation augmentation, and EMA. The model takes RGB LF input, reconstructs the luminance component in YCbCr space, and converts the result back to RGB. Unless otherwise stated, all quantitative results for the main model use this setting.

\paragraph{GTF-Tiny.} We use learning rate $4 \times 10^{-4}$ with StepLR (step size 80, $\gamma = 0.5$), batch size 8, and LR patch size $32 \times 32$. Training runs for 180 epochs (about 20 hours on A100) with EMA decay 0.999.

Both models are trained exclusively on the standard NTIRE 2026 training data, without external datasets or pretrained weights.

\subsection{Inference Protocol}

We use \textbf{EPSW}~\cite{jin2025lftransmamba} to aggregate overlapping patches and reduce boundary artifacts. We also apply 8-fold \textbf{TTA} with flips and $90^\circ$ rotations when reporting the strongest Track~1/3 results. Unless otherwise stated, the architecture-evolution, channel-width, and FFN analyses are reported without EPSW or TTA to isolate model effects. For the final Track~1 and Track~3 submissions, we further average the outputs of GTF, LF\_DET~\cite{cong2023lfdet}, and LFTransMamba~\cite{jin2025lftransmamba} with empirically selected weights.

\section{Experiments}
\label{sec:experiments}

\subsection{Experimental Setup}
\label{sec:setup}

We evaluate the proposed models on five standard LF benchmarks widely used in the literature~\cite{wang2023ntire,NTIRE2026-LFSR}: EPFL, HCI\_new, HCI\_old, INRIA\_Lytro, and Stanford\_Gantry. EPFL, INRIA\_Lytro, and Stanford\_Gantry are real-captured, whereas HCI\_new and HCI\_old are synthetic. The task is $4\times$ spatial super-resolution with $5\times5$ angular resolution, and we report PSNR/SSIM on the Y channel under the standard protocol.

Our models are trained on the official NTIRE 2026 training set without external data or pretrained weights. The detailed optimization and inference settings are given in Sec.~\ref{sec:training}.

\subsection{Competition Results}
\label{sec:competition}

This work is part of the NTIRE 2026 Light Field Image Super-Resolution Challenge~\cite{NTIRE2026-LFSR}. We participate in all three tracks and use the challenge as an additional evaluation of the proposed design. For Track~1 and Track~3 (Fidelity), the final submission uses weighted averaging of GTF, LF\_DET~\cite{cong2023lfdet}, and LFTransMamba~\cite{jin2025lftransmamba}, placing 3rd on both tracks. For Track~2 (Efficiency), we submit GTF-Tiny as a single model under the ${<}1$M parameter and ${<}20$G FLOP constraints, placing 4th with 0.915M parameters and 19.81 GFLOPs. The challenge also evaluates hidden extra scenes beyond the five public benchmarks, including both real and synthetic sources. We therefore use the five public benchmarks for detailed method-to-method comparison and treat the official challenge ranking as complementary evidence on the hidden challenge scenes.

\begin{figure*}[!t]
\centering
\includegraphics[width=\linewidth,trim=10 12 10 12,clip]{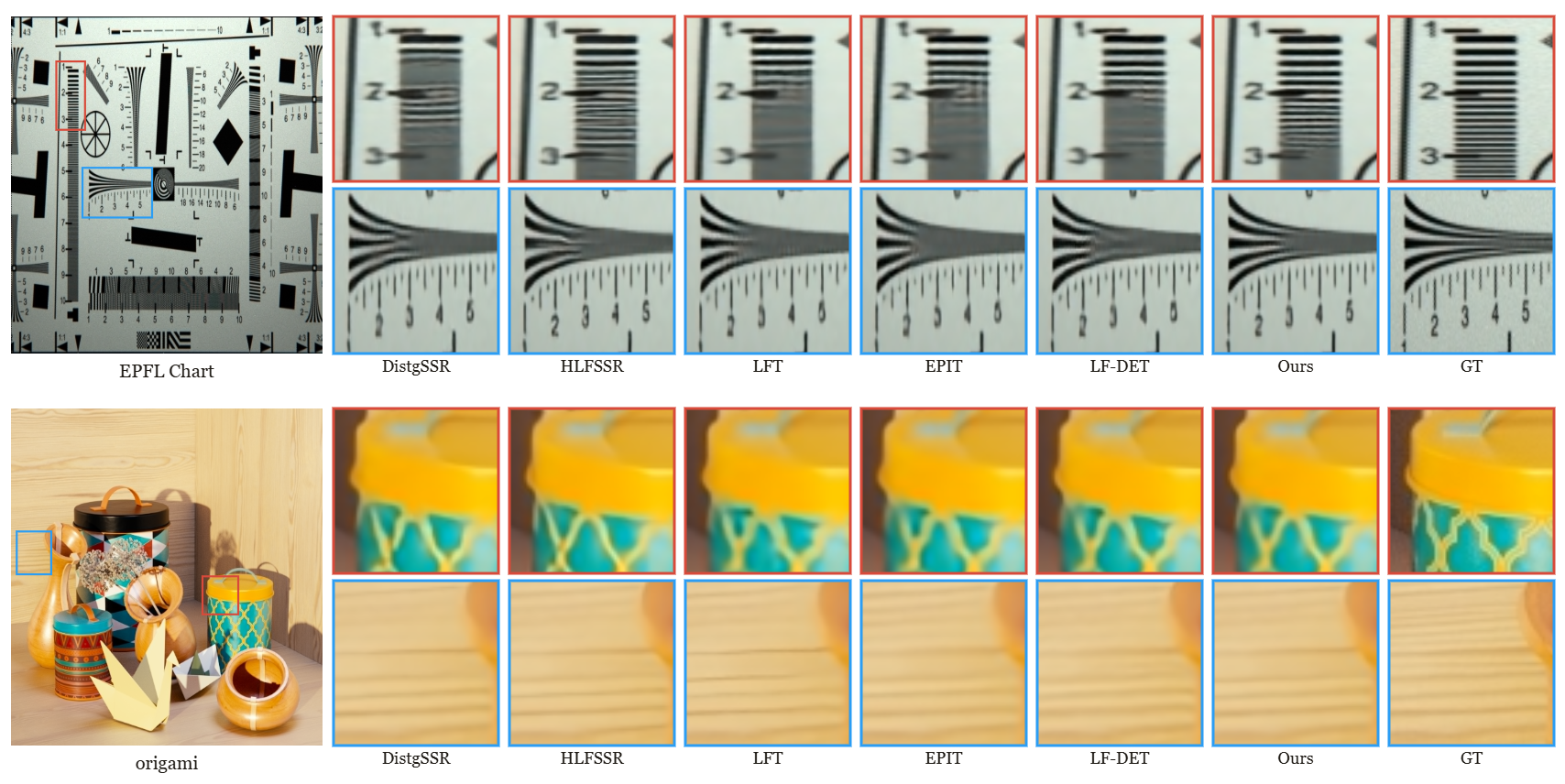}
\caption{Qualitative comparison on representative real and synthetic benchmark scenes. From left to right, each group shows DistgSSR, HLFSR, LFT, EPIT, LF-DET, GTF, and GT on the same crop. Compared with the representative baselines, GTF recovers clearer local structures and cleaner textures with fewer artifacts.}
\label{fig:qualitative}
\end{figure*}

\subsection{Comparison with State-of-the-Art}
\label{sec:sota}

Table~\ref{tab:sota} compares the proposed models with representative prior methods on the five standard benchmarks. Without inference-time enhancement, GTF attains 32.78~dB, outperforming the compared prior methods in Table~\ref{tab:sota}, including LF-DET~\cite{cong2023lfdet} (32.48~dB). With EPSW~\cite{jin2025lftransmamba} and TTA, GTF reaches 33.16~dB. Under the efficiency constraint, GTF-Tiny obtains 32.57~dB with only 0.915M parameters and 19.8 GFLOPs using EPSW only (without TTA), exceeding EPIT~\cite{liang2023epit} (32.40~dB) while using 37\% fewer parameters. These results support the effectiveness of the proposed omnidirectional design on both real-captured and synthetic benchmark sets, while the challenge ranking provides complementary evidence on held-out scenes.

\begin{table}[H]
\centering
\small
\setlength{\tabcolsep}{4pt}
\setlength{\abovecaptionskip}{2pt}
\setlength{\belowcaptionskip}{0pt}
\renewcommand{\arraystretch}{1.06}
\caption{Representative architecture evolution from the EPIT-style baseline to the final reported GTF model. Mean PSNR/SSIM averaged over all five benchmarks.}
\label{tab:ablation_arch}
\begin{tabular}{@{}>{\raggedright\arraybackslash}m{0.29\columnwidth}>{\raggedright\arraybackslash}m{0.33\columnwidth}cc@{}}
\toprule
Configuration & Key Design Change & PSNR & SSIM \\
\midrule
EPIT baseline~\cite{liang2023epit} & --- & 32.40 & 0.9452 \\
+ MacPI Prior + Depth & MacPI prior, 8 blocks & 32.45 & 0.9452 \\
+ Diagonal Branch & Diagonal EPI branch & 32.52 & 0.9454 \\
+ Direction Fusion & Adaptive directional fusion & 32.56 & 0.9458 \\
+ TP-FFN & Topology-preserving FFN & \textbf{32.65} & \textbf{0.9470} \\
\bottomrule
\end{tabular}
\end{table}

\subsection{Ablation Studies}
\label{sec:ablation}

Unless otherwise stated, the following analyses are conducted without EPSW or TTA so that the measured gains reflect architectural and training effects rather than inference-time enhancement.

\noindent\textbf{Architecture evolution.}
Table~\ref{tab:ablation_arch} summarizes the main architecture evolution from the EPIT-style starting point to the final reported GTF model. This table should be interpreted as a milestone comparison of representative design stages under a unified evaluation protocol, rather than as a strict one-factor controlled ablation. The results show that MacPI prior injection with a deeper backbone, explicit diagonal modeling, directional fusion, and TP-FFN all contribute to the final GTF performance.

\noindent\textbf{Channel-width and FFN analysis.}
Table~\ref{tab:ablation_compact} summarizes two compact follow-up analyses after fixing the main architectural components. Increasing channel width consistently improves reconstruction quality, and the 128-channel setting is adopted in the final reported GTF model. Under the same 128-channel configuration, replacing the 1D FFN with TP-FFN yields +0.21~dB, confirming that preserving 2D EPI topology during feed-forward processing is beneficial.

\noindent\textbf{Loss function.}
Under the final GTF training framework, OHEM improves the mean PSNR from 32.65~dB with plain L1 to 32.784~dB, suggesting that emphasizing hard pixels is helpful when reconstruction difficulty varies across structures and views.

\noindent\textbf{Inference strategies.}
Inference-time enhancement provides further gains. From the vanilla setting (32.775~dB), EPSW reaches 33.006~dB and TTA alone reaches 32.896~dB; combining them yields 33.157~dB.

\begin{table}[H]
\vspace{2pt}
\centering
\small
\setlength{\tabcolsep}{4pt}
\setlength{\abovecaptionskip}{2pt}
\setlength{\belowcaptionskip}{0pt}
\renewcommand{\arraystretch}{1.06}
\caption{Compact follow-up analyses after fixing the main architectural components. Top: channel-width comparison. Bottom: FFN design comparison under the 128-channel configuration.}
\label{tab:ablation_compact}
\begin{tabular}{@{}lcc@{}}
\toprule
\multicolumn{3}{@{}l@{}}{\textbf{Channel-width analysis}} \\
\midrule
Channels & PSNR & SSIM \\
\midrule
64 & 32.41 & 0.9449 \\
96 & 32.64 & 0.9471 \\
128 & \textbf{32.78} & \textbf{0.9480} \\
\midrule
\multicolumn{3}{@{}l@{}}{\textbf{FFN design}} \\
\midrule
FFN Type & PSNR & Note \\
\midrule
1D FFN (EPIT original) & 32.57 & No 2D topology awareness \\
TP-FFN & \textbf{32.78} & 2D DWConv + GELU \\
\bottomrule
\end{tabular}
\end{table}

\vspace{-4pt}
\subsection{Qualitative Results}
\label{sec:qualitative}

Fig.~\ref{fig:qualitative} presents qualitative comparisons on one real-captured scene (EPFL Chart) and one synthetic scene (origami) under a multi-method comparison setting. Relative to representative baselines including DistgSSR~\cite{wang2022distgssr}, HLFSR~\cite{duong2023hlfsr}, LFT~\cite{wang2023lft}, EPIT~\cite{liang2023epit}, and LF-DET~\cite{cong2023lfdet}, GTF recovers clearer local structures and more regular textures with fewer visible artifacts. In the EPFL Chart scene, GTF preserves the thin stripe pattern and local spacing more faithfully, whereas several competing methods produce blur or irregular line continuity around the cropped ruler region. In the origami scene, GTF yields cleaner boundaries around the colored object contours and suppresses over-smoothed transitions in the repeated texture area. All methods are compared at the same crop positions for each scene. These visual results are intended to illustrate reconstruction quality on representative scenes; the role of diagonal EPI modeling is more directly supported by the architecture-evolution analysis in Sec.~\ref{sec:ablation} and the method discussion in Sec.~\ref{sec:diabranch}.

\section{Conclusion}
\label{sec:conclusion}

We presented GTF, an omnidirectional EPI Transformer for light field image super-resolution. By extending Transformer-based EPI modeling beyond the standard horizontal and vertical directions, GTF introduces diagonal epipolar attention and adaptive directional fusion within a unified reconstruction framework. Together with MacPI prior injection in the fidelity setting and the topology-preserving FFN, this design improves the use of LF geometry in attention-based LF reconstruction. Experiments on standard real-captured and synthetic benchmarks, architecture-evolution and channel-width analyses, and the NTIRE 2026 Light Field Image Super-Resolution Challenge support the effectiveness of the proposed design in both fidelity-oriented and efficiency-oriented settings. In particular, GTF is competitive in the fidelity-oriented setting, while GTF-Tiny retains four-direction epipolar reasoning under strict resource constraints. A natural next step is to improve generalization on more challenging held-out LF scenes and to explore stronger efficiency-performance trade-offs for larger angular resolutions and broader LF restoration tasks.

{
    \FloatBarrier
    \small
    \bibliographystyle{ieeenat_fullname}
    \bibliography{main}
}

\end{document}